\documentclass[journal]{IEEEtran}
\usepackage{amsmath,amsfonts}
\usepackage{algorithmic}
\usepackage{algorithm}
\usepackage{array}
\usepackage[caption=false,font=normalsize,labelfont=sf,textfont=sf]{subfig}
\usepackage{textcomp}
\usepackage{multirow}
\usepackage{stfloats}
\usepackage{url}
\usepackage{verbatim}
\usepackage{graphicx}
\usepackage{cite}
\usepackage{colortbl}
\usepackage{color}
\usepackage{xcolor}
\usepackage{makecell}
\usepackage{booktabs}
\usepackage{arydshln}
\usepackage[breaklinks,colorlinks]{hyperref}
\usepackage{soul}
\newcommand{\PreserveBackslash}[1]{\let\temp=\\#1\let\\=\temp}

\newcolumntype{C}[1]{>{\PreserveBackslash\centering}p{#1}}
\newcolumntype{R}[1]{>{\PreserveBackslash\raggedleft}p{#1}}
\newcolumntype{L}[1]{>{\PreserveBackslash\raggedright}p{#1}}

\begin{document}
\title{Depth-Assisted Network for Indiscernible Marine Object Counting with Adaptive Motion-Differentiated Feature Encoding}

\author{Chengzhi~Ma,
        Kunqian~Li,~\IEEEmembership{Member,~IEEE},
        Shuaixin~Liu,
        and Han Mei
        % <-this % stops a space
\thanks{The research has been supported by the National Natural Science Foundation of China under Grant 62371431 and 61906177, in part by the Marine Industry Key Technology Research and Industrialization Demonstration Project of Qingdao under Grant 23-1-3-hygg-20-hy, and in part by the Fundamental Research Funds for the Central Universities under Grants 202262004. (Corresponding author: Kunqian Li)}% <-this % stops a space

\thanks{Chengzhi Ma, Kunqian Li, Shuaixin Liu, and Han Mei are with the College of Engineering, Ocean University of China, Qingdao 266404, China (machengzhi@stu.ouc.edu.cn;  likunqian@ouc.edu.cn; liushuaixin@stu.ouc.edu.cn; meihan@stu.ouc.edu.cn).}% <-this % stops a space

}

\markboth{}%
{Shell \MakeLowercase{\textit{et al.}}: Bare Dem of IEEEtran.cls for Journals}
% *** name in the headers of peer review papers.                  ***



\maketitle


\begin{abstract}
Indiscernible marine object counting encounters numerous challenges, including limited visibility in underwater scenes, mutual occlusion and overlap among objects, and the dynamic similarity in appearance, color, and texture between the background and foreground. These factors significantly complicate the counting process. To address the scarcity of video-based indiscernible object counting datasets, we have developed a novel dataset comprising $50$ videos, from which approximately $800$ frames have been extracted and annotated with around $40,800$ point-wise object labels. This dataset accurately represents real underwater environments where indiscernible marine objects are intricately integrated with their surroundings, thereby comprehensively illustrating the aforementioned challenges in object counting. To address these challenges, we propose a depth-assisted network with adaptive motion-differentiated feature encoding. The network consists of a backbone encoding module and three branches: a depth-assisting branch, a density estimation branch, and a motion weight generation branch. Depth-aware features extracted by the depth-assisting branch are enhanced via a depth-enhanced encoder to improve object representation. Meanwhile, weights from the motion weight generation branch refine multi-scale perception features in the adaptive flow estimation module. Experimental results demonstrate that our method not only achieves state-of-the-art performance on the proposed dataset but also yields competitive results on three additional video-based crowd counting datasets. The pre-trained model, code, and dataset are publicly available at 
\url{https://github.com/OUCVisionGroup/VIMOC-Net}.
\end{abstract}


% \fi
% \IEEEpeerreviewmaketitle
\section{Introduction}
% \IEEEPARstart{S}{ea}
\IEEEPARstart Dense marine biological aggregations are crucial components of marine ecosystems, with fish schools serving as quintessential examples. These densely packed targets exhibit pronounced visual similarity and are visually indistinguishable, making them representative of indiscernible marine objects. Precisely quantifying their abundance and spatial distribution provides researchers with critical insights into their ecological functions and migratory behaviors. Accurate enumeration of indiscernible marine objects is essential to develop more effective strategies for marine conservation and resource management \cite{sun2023indiscernible}. 

However, in underwater environments, light refraction and scattering reduce the visual quality of imaging \cite{Li2024TCTL}, presenting substantial challenges for the observation and enumeration of indiscernible marine objects. Marine objects often aggregate closely, resulting in occlusion and overlap, further complicating the counting process. Moreover, the apparent scale of marine objects is dynamically changed by both their variable swimming speeds and fluctuating distances from the camera, altering perceived size and morphology. Collectively, these factors significantly increase the complexity of accurately counting indiscernible marine objects.

Furthermore, underwater environments frequently feature complex backgrounds \cite{Qi2022SGUIE}. Background elements often exhibit similar appearance, color, and texture to the marine objects in the foreground, particularly under poor lighting conditions, leading to potential confusion between background and foreground objects. Consequently, distinguishing foreground objects becomes more challenging, thus complicating the accurate counting of indiscernible marine objects. Therefore,  counting objects in underwater scenes necessitates not only addressing challenges associated with motion but also effectively differentiating foreground objects from the background.

Previous research on object counting has focused mainly on crowd counting \cite{zhang2016single, li2018csrnet, idrees2018composition, liu2019context, wang2020nwpu, sindagi2020jhu, lin2022boosting, Fan2025Learning}, vehicle counting \cite{onoro2016towards, hsieh2017drone, zhao2022vehicle, Shen2025Lightweight}, and plant counting \cite{lu2017tasselnet, zheng2023multiscale, veramendi2024method}. However, existing approaches face significant challenges when applied to the unique task of indiscernible object counting in marine environments. The IOCFormer \cite{sun2023indiscernible} was developed to address these challenges by integrating density-based and regression-based strategies into a unified framework. Although it is designed to count indiscernible objects within single images in static scenarios, extending its application to video-based counting remains an open challenge. Compared to single-image object counting, video provides continuous spatio-temporal information, which enhances the potential for improving the robustness and accuracy of indiscernible object counting. However, existing video counting methods have predominantly focused on crowd counting tasks \cite{xiong2017spatiotemporal, fang2019locality, hossain2020video, liu2021counting, wu2022spatial, hou2023frame, ling2023motional}, but can hardly handle the distinct challenges associated with counting objects in complex and indiscernible underwater environments, including far more intricate backgrounds, indistinguishable individual appearance differences, and variable movement speeds across diverse scenes. Consequently, this paper addresses the task of counting objects in indiscernible underwater video scenes.

\begin{figure*}[t]
\centering
\includegraphics[width=1\linewidth]{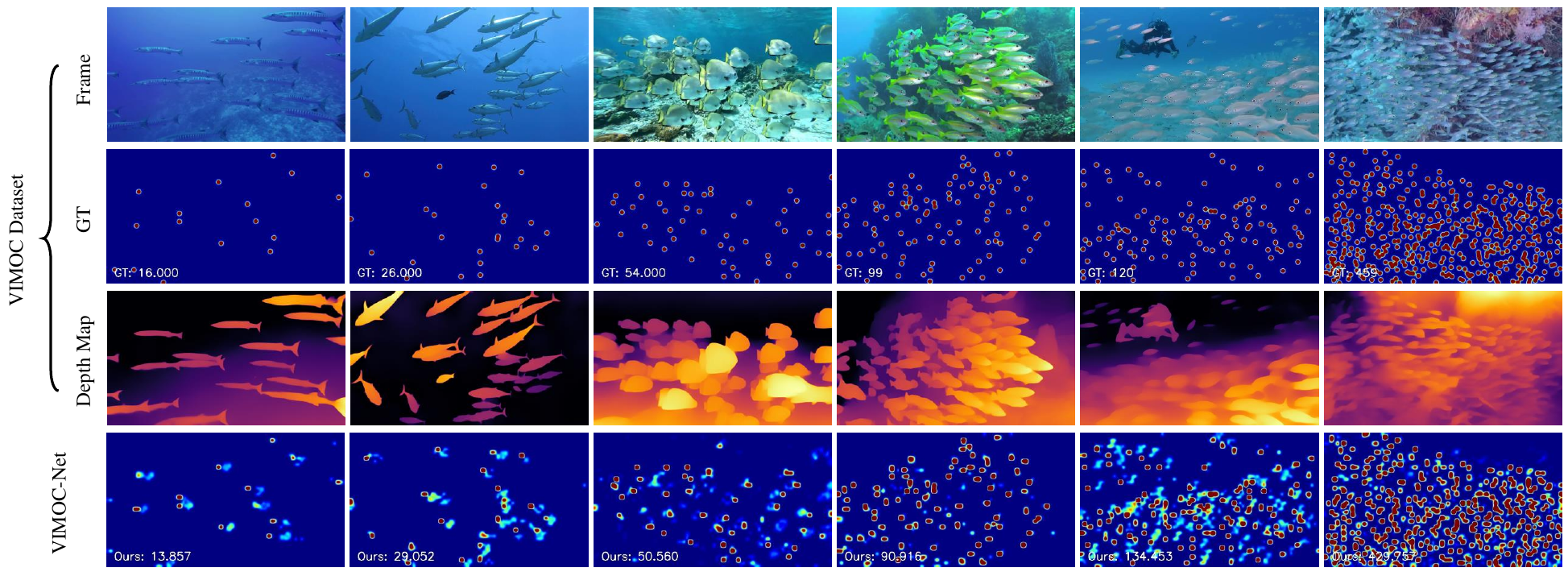}
\caption{Example of the proposed Video Indiscernible Marine Object Counting dataset (VIMOC dataset) and the prediction of our VIMOC-Net. From top to bottom: sampling frames, the corresponding pixel markers of indiscernible objects, the depth maps estimated with the Depth Anything \cite{yang2024depth}, and our prediction.}
\label{fig:dataset}
\end{figure*}

In this paper, to address the novel challenges and opportunities in video-based indiscernible object counting, we introduce a depth-assisted network with adaptive motion-differentiated feature encoding, termed VIMOC-Net, which aims to highlight the individual appearance difference of marine indiscernible objects and better describe them with motion cues. Beyond that, to further facilitate the study of this novel task, we have established a new video-based  indiscernible marine object counting dataset (VIMOC Dataset). This dataset comprises $50$ high-definition fish videos, with each 10th frame annotated, resulting in approximately $40,800$ annotated points in total. In addition, we utilized the currently advanced depth estimation method - Depth Anything \cite{yang2024depth}, to provide single-frame relative depth maps for the dataset, facilitating the research on object counting under a multi-task learning framework guided by depth estimation tasks. Illustrations of the newly established VIMOC dataset and predictions generated by the proposed VIMOC-Net are presented in Fig. \ref{fig:dataset}.
The primary contributions of this paper are as follows:

• A depth-assisted network with adaptive motion-differentiated feature encoding is proposed for indiscernible marine object video counting. To enhance the spatial feature representation of indiscernible objects, we design a depth-enhanced encoder that leverages depth-aware features to refine the regions containing such objects.

• An adaptive flow estimation module is designed that employs motion weights to perform adaptive flow estimation on multi-scale perception features, effectively addressing the issue of motion scale variations caused by differing indiscernible object movement speeds in underwater scenes.

• We establish a new video-based underwater indiscernible object counting dataset, containing $50$ videos, $800$ images, and $40,800$ accurately annotated points. We compare our method with $11$ mainstream counting models and the results show that our method outperforms all $11$ models on this dataset.

%%Since the only publicly available dataset for image-based underwater indiscernible object counting is the IOCfish5K \cite{sun2023indiscernible}, and video-based underwater indiscernible object counting dataset, YoutubeFish-35 \cite{yang2024density}, is not publicly available. Therefore, we establish a new video-based underwater indiscernible object counting dataset, which contains 50 high-definition fish videos, each recorded at 30 frames per second. To enrich the data sample, we annotate every 10th frame in the video using point, resulting in approximately 40,800 annotated points in total. Sample images are shown in Fig. \ref{fig:dataset}. Depth maps are generated by large model Depth Anything \cite{yang2024depth} that is trained using vast data sets to perform well in multiple scenes, including underwater environments. Our newly established video dataset includes various fish species and different swarm sizes, with fish in the video highly integrated with the surrounding environment. This effectively reflects challenging, indiscernible characteristics in underwater scenes, allowing for more efficient performance evaluation. Based on our video counting dataset, we compare our method with 11 mainstream crowd counting methods. Experimental results show that our method significantly outperforms others in accuracy. 

\section{Related Works}\label{sec:Rel}
\IEEEpubidadjcol 
\subsection{Image-based Object Counting}
In the field of image-based object counting, current research predominantly focuses on crowd counting. Notable benchmarks in this domain include ShanghaiTech \cite{zhang2016single}, UCF-QNRF \cite{idrees2018composition}, JHU-CROWD \cite{sindagi2020jhu}, and NWPU-Crowd \cite{wang2020nwpu}. The existing image-based object counting methods can be broadly categorized into three main approaches based on their counting strategies: detection, regression, and density map generation. 

The detection-based object counting method \cite{liu2018decidenet, liu2019point, sam2020locate, wu2023boosting}, which identifies individual objects before counting, generally performs well in sparse scenes. However, accuracy can be significantly compromised in dense scenes due to occlusion and overlap among objects, leading to enumeration errors of individual entities. In contrast, regression-based counting methods \cite{song2021rethinking, liang2022end} directly predict the coordinates of target objects within an image and employ deep learning architectures to establish a mapping between image features and object counts. While this approach can partially alleviate the occlusion issue in dense scenes, its counting accuracy may remain insufficient when handling tightly packed objects, such as the indiscernible marine objects on which this paper focuses. The density map-based approach \cite{zhang2016single, li2018csrnet, lin2022boosting, han2023steerer} generates a density map to estimate the spatial distribution of objects and subsequently counts them by predicting the target density. For example, IOCFormer \cite{sun2023indiscernible} integrates density-based and regression-based counting methodologies within a unified framework, leveraging the advantages of both approaches to enhance the accuracy of indiscernible object counting in underwater environments. Meanwhile, CrowdDiff \cite{ranasinghe2024crowddiff} uses a diffusion model to generate high-fidelity density maps, although this approach requires a comparatively longer training period. In addition to the aforementioned approaches, another category of methods \cite{ma2019bayesian, wang2020distribution, wan2021generalized, yan2023progressive} improves the accuracy of the counting by using improved loss functions.

However, a single image provides only static information about the scene and objects of interest. Precisely localizing densely aggregated and indiscernible objects, such as highly occluded and densely packed fish in marine environments, within this static context poses significant challenges. 

\subsection{Video-based Object Counting}
Videos provide continuous spatio-temporal variation information, where spatial structure cues and motion dynamics within frames offer significant potential to improve the accuracy and robustness of counting algorithms. The core issues focused by existing methods include the acquisition of density maps and the methodologies for feature association and fusion under spatio-temporal constraints. 

As an early attempt, ConvLSTM \cite{xiong2017spatiotemporal} uses a bidirectional convolutional LSTM to capture long range temporal information. LSTN \cite{fang2019locality} first uses a convolutional neural network to estimate the density map for each frame, and then uses a spatial transformer network (STN) \cite{jaderberg2015spatial} to model the relationship between the density maps of adjacent frames. MOPN \cite{hossain2020video} integrates an optical flow pyramid with an pre-trained feature extractor, incorporating motion information via multi-scale embeddings derived from optical flow for the joint prediction of density maps. EPF \cite{liu2021counting} proposes a flow-based method to estimate crowd density in video sequences, advocating that the flow of people across image locations should be estimated between successive frames, with the flow density inferred from these image streams instead of using direct regression. STGN \cite{wu2022spatial} creatively estimates the density of the crowd in videos by modeling the relationship between pixels and patches within the local spatio-temporal domain through the introduction of a pyramid graph module. FRVCC \cite{hou2023frame} introduces a frame-recurrent mechanism that recursively correlates density maps across the temporal dimension, thereby effectively leveraging long-term interframe information and ensuring the temporal consistency of feature map responses. MAF \cite{ling2023motional} introduces moving foreground attention, which extracts motion features through two-way frame differencing and combines these motion features with the static features of the current frame to estimate the crowd count.

However, the existing works primarily address crowd counting in videos, the challenge of counting indiscernible objects in marine environments presents more formidable obstacles, including complex backgrounds, indistinguishable target appearances, and varying movement speeds across different scenes. In this paper, we utilize depth-aware features to enhance the differentiation of indiscernible objects. Furthermore, we propose adaptively encoding motion-differentiated features using flow guidance, thereby enhancing the learning of more reliable and reasonable features across varying motion rates.

\subsection{Visual Object Counting with Auxiliary Data}
In visual counting tasks, in addition to fully exploiting information from RGB images, a prevalent strategy is to integrate additional auxiliary data, such as depth maps and thermal, to provide additional cues. This has made RGB-Depth (RGB-D) and RGB-Thermal (RGB-T) object counting particularly active research areas. For example, Lian et al. \cite{lian2019density} proposed the RDNet model, which estimates crowd head counts using RGB-D information while simultaneously locating heads with bounding boxes. The depth-adaptive kernel was introduced for more robust density map regression. Liu et al. \cite{liu2021cross} proposed a cross-modal collaborative representation learning framework aimed at comprehensively capturing complementary information across different modalities, thus improving counting performance. TAFNet \cite{tang2022tafnet} is a novel three-stream network specifically designed for RGB-T crowd counting, which adaptively integrates RGB and thermal images, thus improving adaptability and achieving higher accuracy in complex environments. More recently, CSA-Net \cite{li2023csa} introduced a scale-sensing cross-modal feature aggregation module and a scale-aware channel attention aggregation module to improve the performance of RGB-T crowd counting. Meng et al. \cite{meng2025multi} introduced a novel triple-modal learning framework for multi-modal crowd counting, which incorporates an auxiliary broker modality generated via a lightweight fusion model. 

However, we observe that all of the aforementioned schemes require auxiliary data as input for model prediction, which is impractical in numerous scenarios and incurs additional application costs. In this paper, we concentrate on integrating auxiliary data as supplementary supervision during the model training phase, rather than employing them directly as part of the model input. This approach helps to better adapt it to video tasks and expand the application scenarios of the algorithm.

\section{Proposed Method}\label{sec:Pro}
Similar to the concept of terrestrial video crowd counting \cite{wu2022spatial, hou2023frame, ling2023motional}, which builds upon the task of image object counting, we define underwater video indiscernible object counting as the estimation of the number of indiscernible underwater objects in a single frame, utilizing information from consecutive frames within the video sequence. As a novel attempt for the above new challenging task, we introduce VIMOC-Net, a depth-assisted network that incorporates adaptive motion-differentiated feature encoding to address the challenges associated with indiscernible object counting in underwater videos. In this section, we first provide an architectural overview and analysis of VIMOC-Net, followed by a detailed description of its key modules.

\begin{figure*}[t]
  \centering
   \includegraphics[width=0.98\linewidth]{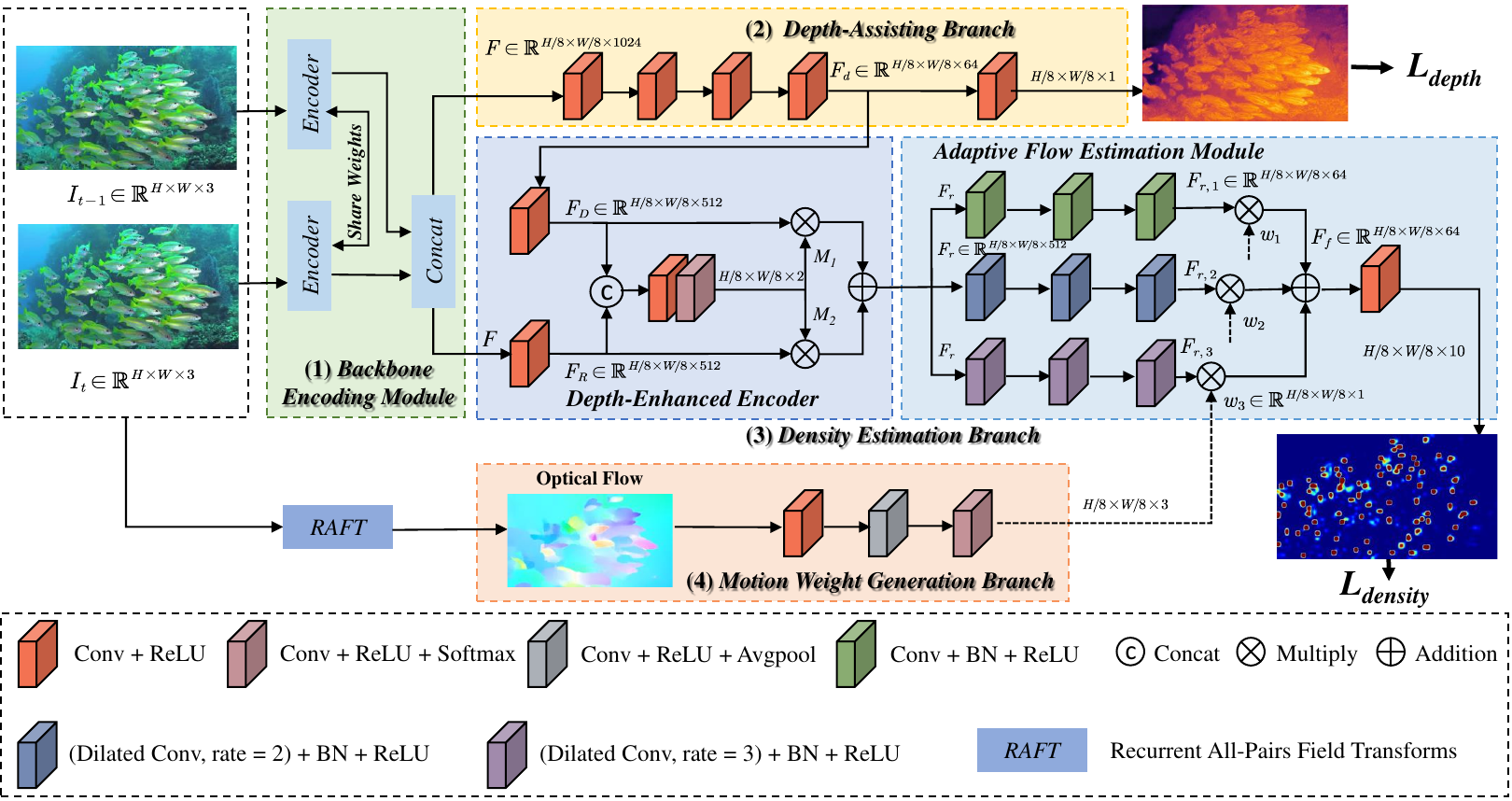}
   \caption{The framework of the proposed VIMOC-Net, which is a depth-assisted indiscernible marine object counting network with adaptive motion-differentiated feature encoding. The network input consists of the previous frame $I_{t-1}$ and current frame $I_t$. After feature extraction by a shared encoder, the features from both frames are concatenated into a feature map $F$, which is then fed into the depth-assisting branch and density estimation branch. The depth-enhanced encoder leverages depth-aware features $F_d$ to improve indiscernible object features. The adaptive flow estimation module applies motion weights $w_i$ to estimate flow adaptively on multi-scale perception features.}
   \label{fig:network}
\end{figure*}

\subsection{Architectural Overview and Analysis of VIMOC-Net}
The overall framework of VIMOC-Net is shown in Fig. \ref{fig:network}. It comprises a backbone encoding module and three branches: the density estimation branch, the depth-assisting branch, and the motion weight generation branch. The depth-assisting and motion weight generation branches serve as auxiliary components that enhance the performance of the density estimation branch by providing spatio-temporal guidance, thereby improving the accuracy of predictions for indiscernible objects in complex underwater environments where static information alone is inadequate.

\subsubsection{Backbone Encoding Module}
We leverage the first 13 layers of VGG16 \cite{simonyan2014very} and the Contextual Module from CAN \cite{liu2019context} to construct the backbone encoding module. Both the preceding frame $I_{t-1}\in \mathbb{R} ^{H \times W \times 3}$ and the current frame $I_{t}\in \mathbb{R} ^{H \times W \times 3}$ are fed into the encoder, which extracts high-dimensional features from each frame, reducing the spatial dimensions to one-eighth of their original size. The extracted features from the two frames are then concatenated to form a fused feature map $F\in \mathbb{R} ^{H/8 \times W/8 \times 1024}$, which is subsequently processed by both the depth-assisting branch and the depth-enhanced encoder (DEE) to facilitate advanced perception and transformation.

\subsubsection{Depth-Assisting Branch}\label{sec:DAB}
The depth-assisting branch is designed to leverage a joint-learning depth estimation sub-task to learn feature representations that incorporate image depth perception. These learned auxiliary features are subsequently integrated into the main branch, the density estimation branch, to improve its performance. The depth-assisting branch processes the feature map $F$ through a depth decoder composed of four convolutional layers, each equipped with 3$\times$3 kernels and ReLU activation functions. The resulting output features are denoted as $F_d \in \mathbb{R} ^{H/8 \times W/8 \times 64}$. Thereafter, the branch head transforms $F_d$ into a non-negative, single-channel depth map $D$, which is supervised by the pseudo ground truth depth map generated using Depth Anything \cite{yang2024depth}. The depth loss $L_{depth}$ is calculated as the mean squared error (MSE) between the predicted depth map $D$ and the corresponding ground truth depth reference. As derived from the intermediate layers of the depth-assisting branch, the feature map $F_d$ can be regarded as a depth-aware representation that encapsulates rich spatial information to help distinguish indiscernible object instances.

In this depth-assisting branch, given the computational and fine-tuning costs associated with the main branch task, we opted not to directly adopt Depth Anything as the branch architecture. Instead, we employed a knowledge distillation approach, retraining with the generated predictions from Depth Anything. This method provides an effective, low-cost solution aimed at obtaining enough robust representations of auxiliary features for the counting task.

\subsubsection{Density Estimation Branch}
The density estimation branch constitutes the central component of VIMOC-Net. This branch integrates auxiliary information from both the depth-assisting branch and the motion weight generation branch to achieve multi-scale spatio-temporal feature fusion, ultimately generating a density map and estimating the quantity of indiscernible marine objects. 

Specifically, in the depth-enhanced encoder, designed to generate feature representations of counting targets with improved spatial perception capabilities, the feature map $F$ from the backbone encoding module and the depth-aware feature map $ F_{d} $ from the depth-assisting branch are fused to obtain refined features $F_{r}\in \mathbb{R} ^{H/8 \times W/8 \times 512}$. A more comprehensive description of the fusion procedure within the depth-enhanced encoder is provided in Section~\ref{sec:DEE}. 

To further enhance the feature representation's capability to characterize multiple motion scales for density estimation, we developed an adaptive flow estimation module. This module consists of three groups of parallel stacked dilated convolution layers on multiple scales with $F_{r}$ as input, which allows more robust and versatile feature extraction. To further adaptively learn and predict the optimal weights for each group, we incorporate the output of the motion weight generation branch as the predicted weights. Subsequently, multi-scale perception features extracted from various dilated convolution layers are adaptively fused. Ultimately, the density flow prediction is generated through the density head. Detailed information is provided in Section~\ref{sec:adaflow}. 

\subsubsection{Motion Weight Generation Branch}
As mentioned earlier, while multi-scale perception features can be extracted via multi-parameter dilated convolutions, it is crucial to appropriately fuse these features in a differentiated manner by incorporating the motion scale information from the current frame. Given the significance of optical flow as a critical motion perception technique for video analysis, we have developed a mechanism that leverages optical flow transformation to automatically predict weights for multi-scale perception features. The optical flow between the preceding frame $ I_{t-1} $ and the current frame $ I_{t} $ is calculated utilizing the RAFT model \cite{teed2020raft}. The resultant optical flow is subsequently fed into the Motion Weight Generation (MWG) branch to obtain the motion weights $ w_{i} \in \mathbb{R} ^{H/8 \times W/8 \times 1}$, where $i = 1, 2, 3$. This process is detailed in Section~\ref{sec:adaflow}.

\subsection{Depth-Enhanced Encoder for Spatial Information Fusion}
\label{sec:DEE}

The depth-enhanced encoder (DEE) leverages depth-aware features $F_d$ to integrate with the encoder's feature mapping $F$, thereby enhancing the spatial identifiability of indiscernible objects and improving the robustness of feature representation in complex underwater environments. 

\begin{figure}[t]
  \centering
   \includegraphics[width=1\linewidth]{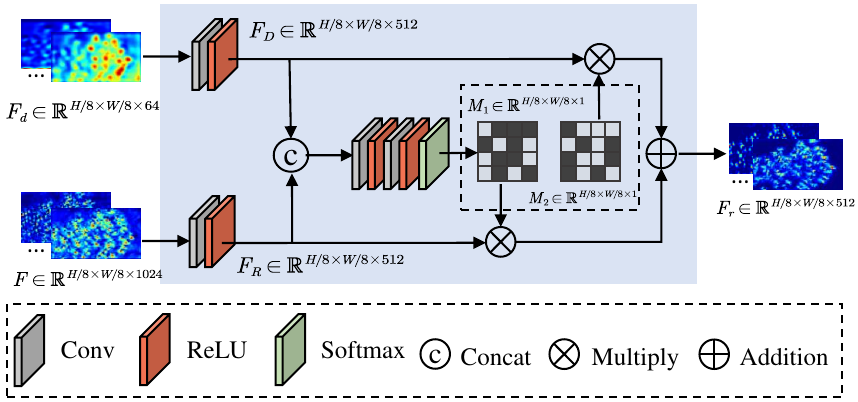}
   \caption{Detailed structure of the Depth-Enhanced Encoder (DEE).}
   \label{fig:DEE}
\end{figure}

Fig. \ref{fig:DEE} delineates the intricate architecture of the depth-enhanced encoder. Initially, DEE employs two parallel convolutional blocks to project the depth-aware features $F_{d}$ and the backbone encoding feature map $F$ into a 512-channel space, ensuring channel alignment for subsequent addition fusion. To enhance the integration of spatial perception from $F_{d}$ and appearance perception from $F$, we introduce an element-wise fusion weight learning strategy. Specifically, the aligned depth features $F_{D}\in \mathbb{R} ^{H/8 \times W/8 \times 512}$ are concatenated with the transformed backbone encoding features $F_{R} \in \mathbb{R} ^{H/8 \times W/8 \times 512}$. This concatenated feature set is then processed through two Conv-ReLU layers to learn joint feature representations. Subsequently, these learned features pass through the Softmax activation function to generate normalized element-wise fusion weights $M_1 \in \mathbb{R} ^{H/8 \times W/8 \times 1}$ and $M_2\in \mathbb{R} ^{H/8 \times W/8 \times 1}$, which are applied for $F_{R}$ and $F_{D}$, respectively. This process ultimately yields the refined features $F_{r}\in \mathbb{R} ^{H/8 \times W/8 \times 512}$ as follows
\begin{equation}
\label{deqn_ex1}
F_r = M_1(F_R,F_D)\otimes F_R +  M_2(F_R,F_D)\otimes F_D,
\end{equation}
where $\otimes$ denotes element-wise multiplication.

We can notice that, by effectively capturing depth information, the model can more accurately differentiate between foreground and background elements. Moreover, these features help identify and improve visually ambiguous target objects, leading to more detailed and precise feature representation. This capability strengthens the predictive performance of the density estimation branch, enabling a more accurate estimation of object density distributions, especially in scenes with complex backgrounds and similar textures. Consequently, the model not only helps achieve superior performance in counting tasks, but also provides a robust solution for detecting indiscernible objects in complex underwater scenes.

\subsection{Motion Temporal Perception for Adaptive Flow Estimation}\label{sec:adaflow}
Unlike terrestrial crowd counting tasks, where the movement speed of individuals under a fixed perspective is relatively constant, the velocity of underwater indiscernible objects (such as schools of fish) exhibits significant variability. Furthermore, the continuous motion of the underwater camera during recording introduces additional complexity, resulting in diverse and unpredictable motion scales of the objects that are counted in the video footage. To better perceive the diverse temporal motions of indiscernible objects, an optical flow-based method is employed to dynamically track motion information and highlight changes in movement characteristics. Then, to address the challenge of varying motion scales in underwater object counting, a multi-stream architecture is utilized to capture features at different motion scales. Furthermore, motion weight-based adaptive fusion is applied to emphasize the most relevant features of moving objects.

\begin{figure}[t]
  \centering
   \includegraphics[width=1\linewidth]{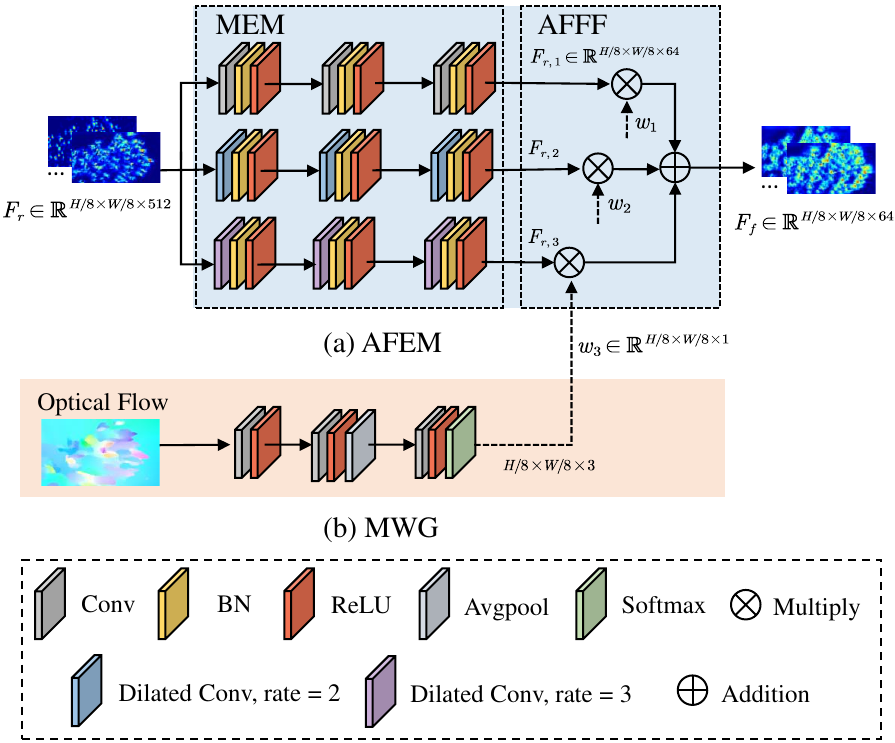}
   \caption{Detailed structures of the Adaptive Flow Estimation Module (AFEM) and the Motion Weight Generation (MWG). The AFEM is composed of the motion enhancement module and the adaptive flow feature fusion.}
   \label{fig:AFEM}
\end{figure}

Fig. \ref{fig:AFEM} presents the detailed architecture of the adaptive flow estimation module and the motion weight generation process. The adaptive flow estimation module comprises two essential components: the motion enhancement module (MEM), which captures features across various motion scales, and the adaptive flow feature fusion mechanism (AFFF), which mitigates environmental interference by accentuating significant features. With $F_r$ as input, the MEM adopts a multi-stream architecture to extract features across various motion scales. Each stream is specifically designed to capture features from different receptive fields by using convolutions with varying dilation rates. More precisely, each stream consists of three convolutional layers, each incorporating a convolution kernel, batch normalization, and a ReLU activation function. The convolution kernels in each stream are distinct: the first stream utilizes standard $3\times3$ convolutions, the second employs dilated $3\times3$ convolutions with a dilation rate of 2, and the third applies dilated $3\times3$ convolutions with a dilation rate of 3. Using dilated convolutions, the receptive field of each layer is significantly enlarged without an increase in the number of parameters. This architectural design allows the model to effectively capture multi-scale perception features while preserving computational efficiency. The specific receptive fields of the motion enhancement module are detailed in Table \ref{tab:table1}.

The features extracted at various motion scales by the MEM are denoted as $ F_{r,i} \in \mathbb{R} ^{H/8 \times W/8 \times 64}$, for $i = 1, 2, 3$. Subsequently, these features are adaptively fused using the motion weights $ w_{i} $. The process of adaptive flow feature fusion is defined as follows:
\begin{equation}
\label{deqn_ex1}
F_{f} = \sum_{i=1,2,3,} w_{i} \otimes F_{r,i}, 
\end{equation}
\begin{equation}
\label{deqn_ex1}
w_{i}  = \frac{e^{w_{i}}}{\sum_{i=1,2,3} e^{w_{i}}}, i=1, 2, 3,
\end{equation}
where $w_i$, for $i = 1, 2, 3$, represents the fusion weights generated by the motion weight generation branch.

Instead of using fixed weight ratios for motion features $F_{r,i}$ during fusion at each scale, we employ adaptive motion weights, which allows the model to dynamically adjust to complex underwater environments with varying target motion speeds. Moreover, by more effectively aligning with the actual data distribution, adaptive fusion not only mitigates overfitting but also enhances model flexibility between training and test sets, thereby improving generalization to new application scenarios. In Section \ref{sec:ablaMWGF}, we perform ablation studies to verify the superiority of motion weight-guided fusion over fixed weight fusion.

\subsection{Object Flow Conservation Constraint}
With the fused feature $F_f \in \mathbb{R}^{H/8 \times W/8 \times 64}$, the density head predicts the density flow for indiscernible marine object counting. Specifically, we integrate the conservation constraint from EPF \cite{liu2021counting} to estimate the density in the current frame $I_t$ by applying a flow prediction strategy between the preceding frame $I_{t-1}$ and the current frame $I_t$. The conservation constraint stipulates that the density at a given position at time $t$ can only originate from adjacent grid positions at time $t-1$ and flow to adjacent grid positions at time $t+1$, including the position itself. This ensures that the model accounts for objects that remain stationary in the same position. Our model outputs a downsampled density map, where each grid position corresponds to an $8\times8$ pixel block in the input image. This downsampling rate is commonly used in crowd-counting models \cite{li2018csrnet}\cite{liu2019context}\cite{liu2021counting}, as it strikes a good balance between the resolution of the density map and the computational efficiency of the model.

We denote the predicted flow direction between all adjacent locations from time $t-1$ to time $t$ using $f^{t-1,t}$. In practice, $f^{t-1,t}$ shares the same dimensions as the image grid, with $10$ channels assigned per location. The first nine channels correspond to the eight neighboring positions that surround a given grid point, along with the position itself. The tenth channel signifies potential flow originating from outside the image boundaries, which holds significance only at the grid's edges. Let $m_{j}^{t}$ represent the number of fish in the grid position $j$ at time $t$, and let $f_{i,j}^{t-1,t}$ denote the number of fish that transitioned from grid position $i$ to the grid position $j$ between times $t-1$ and $t$. For all non-edge positions $j$, the conservation constraint can be formulated as:
\begin{equation}
\label{ex1}
m_{j}^{t} = \sum_{i\in N(j) } f_{i,j}^{t-1,t},
\end{equation}
where $N(j)$ represents the $8$ neighbours surrounding grid location $j$, including $j$ itself, to account for fish that remain in the same location. Let $m_{j}^{t}$ be the aggregate flow of fish in the grid position $j$ at time $t$, then the predicted total count for the entire image can be obtained by summing all $m_{j}^{t}$ values. If the direction of the video sequence is reversed, the flow vector should exhibit the same magnitude but an opposite direction. This relationship can be expressed as:
\begin{equation}
\label{ex2}
f_{i,j}^{t-1,t} =f_{j,i}^{t,t-1}.
\end{equation}

\begin{table}[t]
\caption{The receptive field of the dilation convolution layer.\label{tab:table1}}
\centering
\fontsize{8}{15}\selectfont % 设置字体大小为10pt，行距为15pt
\begin{tabular}{c c c c|c c c|c c c}
\hline
\multicolumn{1}{c|}{Stream}   & \multicolumn{3}{c|}{Top}   & \multicolumn{3}{c|}{Middle} & \multicolumn{3}{c}{Bottom} \\ \hline
\multicolumn{1}{c|}{Layer}   &1   &2  &3   &1 &2 &3    &1 &2 &3 \\  \hline
\multicolumn{1}{c|}{Dilation rate}  & \multicolumn{3}{c|}{1}   & \multicolumn{3}{c|}{2}    & \multicolumn{3}{c}{3} \\\hline
\multicolumn{1}{c|}{\makecell[c]{Receptive field ($n \times n$)}}   &3  &5  &7  &5 &9 &13    &7 &13 &19   \\\hline
\end{tabular}
\end{table}

\subsection{Loss Function}
In line with previous research \cite{zhang2016single, li2018csrnet, liu2019context, liu2021counting}, we employ a similar methodology to generate the ground truth of the density map. Specifically, each image is annotated with $n$ two-dimensional points $\left\{P_{i}^{t}\right\}_{i=1}^{n}$, which indicate the positions of the counting targets within the scene. The corresponding true density map $\hat{m}_{j}^{t}$ is generated through the convolution of binary images, where pixels at specified positions are set to $1$ and all other pixels are set to $0$, with a Gaussian kernel $N(\cdot\mid\mu, \sigma^2)$ characterized by a mean of $\mu$ and a standard deviation of $\sigma$. The true density value for any position $j$ within the image can be written:
\begin{equation}
\label{deqn_ex1}
\hat{m}_{j}^{t} =\sum_{i=1}^{n}N(p_j \mid \mu=P_{i}^{t}, \sigma^2),
\end{equation}
where $p_j$ denotes the center of location $j$, and $\sigma$ is set to $3$ following standard recommendations.

We input the previous frame $I_{t-1}$ and the current frame $I_{t}$ into the network for training to obtain the fish flow estimate. The predicted density $m_{j}^{t}$ of $I_{t}$ is derived from conservation constraints as given in Eq. (\ref{ex1}). To ensure that the predicted density aligns with the actual density, we employ a flow loss function $L_{flow}$ to quantify the discrepancy between the predicted and actual density values, which is expressed as:
\begin{equation}
\label{deqn_ex1}
L_{flow} = \sum_{j \in I_{t}} \left(\hat{m}_{j}^{t} - m_{j}^{t}\right)^{2},
\end{equation}
where $\hat{m}_{j}^{t}$ denotes the actual density value at time $t$ and position $j$, while $m_{j}^{t}$ represents the predicted density value at the same time and position. Then, we enforce the constraint in Eq. (\ref{ex2}) by employing the cycle consistency loss $ L_{cycle} $, formulated as follows:
\begin{equation}
\label{deqn_ex1}
L_{cycle} = \sum_{j \in I_{t}} \sum_{i \in N(j)} \left(f_{i,j}^{t-1,t} - f_{j,i}^{t,t-1}\right)^{2}.
\end{equation}

The density loss $L_{density}$ is computed by summing $L_{flow}$ and $L_{cycle}$. Along with the depth loss $L_{depth}$ of the depth-assisting branch introduced in Section~\ref{sec:DAB}, which utilizes mean square error (MSE) loss, the total loss function for VIMOC-Net is formulated as:
\begin{equation}
\label{deqn_ex1}
L = L_{flow} + L_{cycle} + L_{depth}.
\end{equation}

\begin{figure}[t]
\centering
\includegraphics[width=1\linewidth]{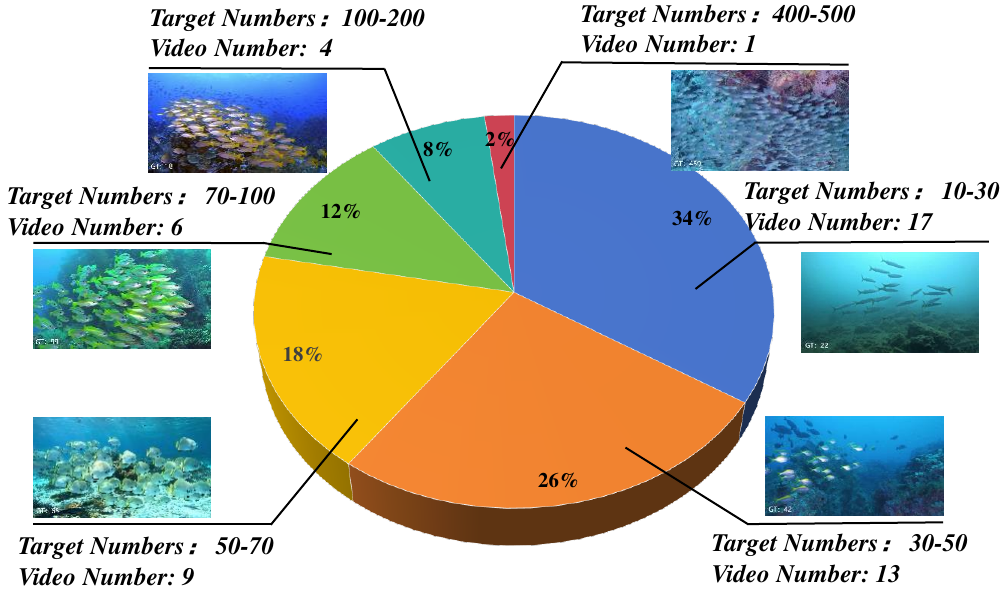}
\caption{The proportion of indiscernible objects within specific number ranges across the entire video dataset. A representative sample from each number range is selected for display, with the corresponding count labeled at the lower left corner of each sample.}
\label{fig:number}
\end{figure}

\section{Experiment}\label{sec:Expe}

\subsection{Implementation Details and Experimental Setting}

\subsubsection{Implementation Details}
During the training process, the Adam optimizer \cite{kingma2014adam} was employed for optimization. The initial learning rate was set to $1 \times 10^{-4}$, with a weight decay of $5 \times 10^{-4}$. A batch size of 1 and input image dimensions of $640 \times 360$ were utilized. All experiments were conducted on a consistent hardware configuration consisting of a single NVIDIA RTX 3090 GPU, a 2.3 GHz Intel Xeon processor, 64 GB RAM, and the Ubuntu 18.04 operating system.

\subsubsection{Dataset}
We have established an indiscernible marine object counting dataset comprising $50$ fish video sequences. The frames were extracted at intervals of every 10 frames, yielding a total of $800$ images annotated with approximately $40,800$ precise points. This dataset encompasses a diverse array of fish species and schooling configurations, ranging from small clusters to densely packed formations, thereby demonstrating a high degree of diversity and complexity. In Fig. \ref{fig:number}, we present the distribution of fish densities along with illustrative examples. Of the total videos, $60\%$ ($30$ out of $50$) contain $0\sim 50$ fish. Additionally, $30\%$ ($15$ out of $50$) of the videos contain $50\sim 100$ fish. Notably, $5$ videos ($10\%$) feature more than $100$ fish, with an exceptional case containing $400\sim 500$ fish. This density distribution of the counting targets offers a basis for evaluating the performance of the indiscernible object counting method across varying densities.

\begin{figure}[t]
\centering
\includegraphics[width=0.98\linewidth]{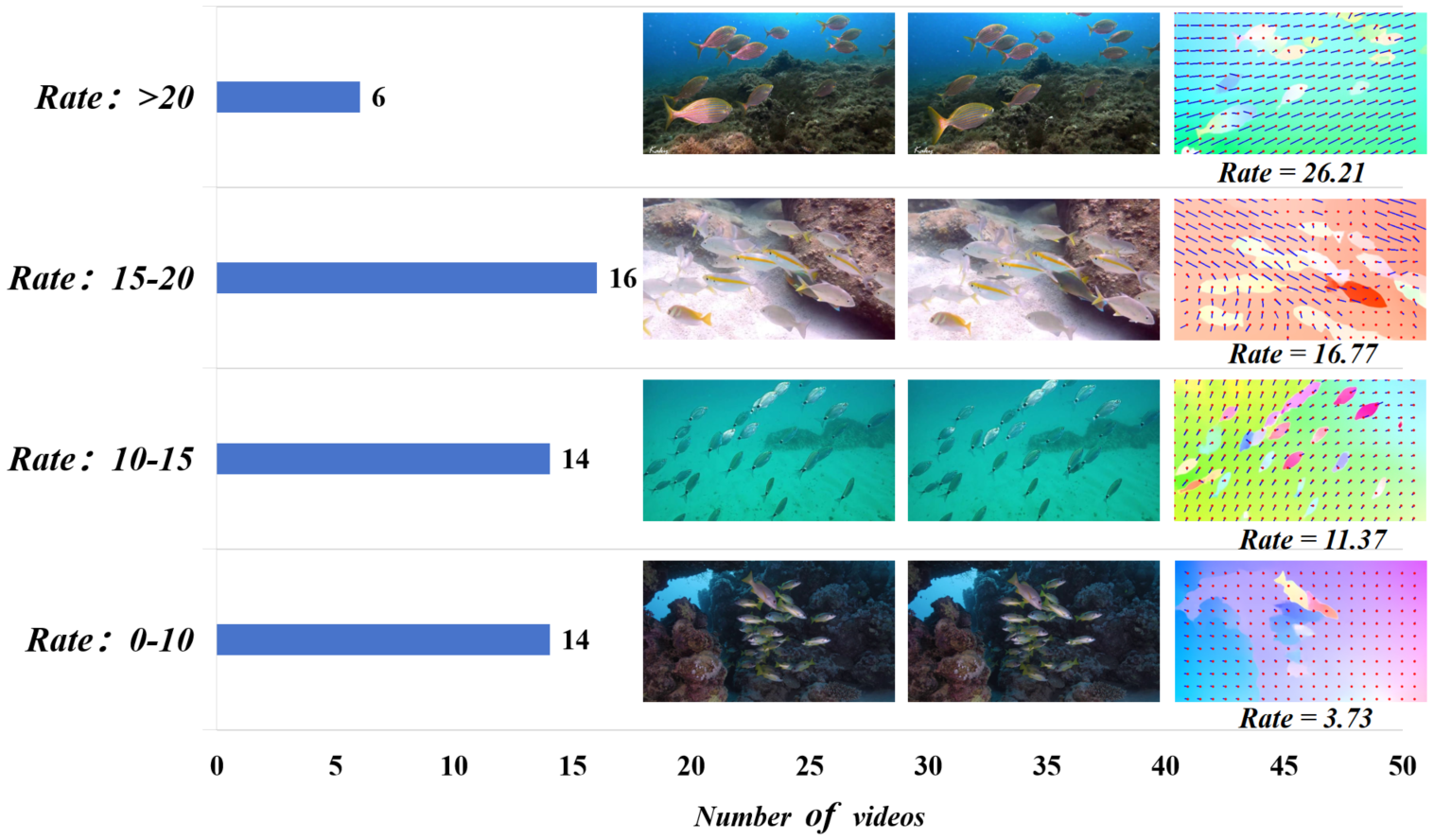}
\caption{The proportion of the motion rate range for indiscernible objects is presented. A representative sample from each rate range is selected and its motion rate is displayed.}
\label{fig:rate}
\end{figure}

Furthermore, we employ optical flow to analyze and categorize the movement velocity of the counting targets. Optical flow comprises horizontal and vertical components, from which we derive the motion rate by integrating these components, which can be expressed as:
\begin{equation}
\label{deqn_ex1}
Rate = \frac{1}{n} \sum_{i=1}^{n}\sqrt{(u_i)^2+(v_i)^2}  
\end{equation}
where $n$ denotes the number of pixels, $u_i$ and $v_i$ represent the horizontal and vertical components of optical flow on pixel $i$, respectively. In Fig. \ref{fig:rate}, the videos are categorized according to their corresponding motion rates. Specifically, scenes exhibiting a motion rate below $15$ are classified as slow-motion scenes, whereas those with a motion rate exceeding $15$ are designated as fast-motion scenes. Within our dataset, there are $28$ videos of slow-motion scenes and $22$ videos of fast-motion scenes. This classification framework provides robust support for evaluating the model's performance in indiscernible object counting across varying motion rates.

To ensure the scientific rigor and effectiveness of model training and evaluation, we designed a series of three cross-validation experiments. In each experiment, the $50$ video sequences were randomly partitioned into a training set and a test set, with $35$ videos allocated for training and the remaining $15$ reserved for testing. Importantly, each experiment utilized a distinct set of $15$ non-overlapping videos as the test set. This splitting strategy not only provides comprehensive support for assessing the model's performance under varying densities and motion speeds but also minimizes biases introduced by differences in data partitioning, thereby enhancing the stability and reliability of the experimental results. Furthermore, substantial variations in fish numbers and dynamic characteristics within the dataset facilitate the evaluation of the model's capability to handle complex scenes, address occlusion challenges, and achieve accurate counting.

\subsubsection{Evaluation metrics}
To assess the counting performance of the proposed method, we employ mean absolute error (MAE) and root mean square error (RMSE) as evaluation metrics, i.e.,
\begin{equation}
\label{deqn_ex1}
MAE = \frac{1}{n} \sum_{i=1}^{n} |y_{i}-\hat{y_{i}}| ,
\end{equation}

\begin{equation}
\label{deqn_ex1}
RMSE = \sqrt{\frac{1}{n} \sum_{i=1}^{n} (y_{i} - \hat{y}_i)^2} ,
\end{equation}\\
where $n$ denotes the number of test sample frames, $y_{i}$ represents the true value, and $\hat{y_{i}}$ indicates the predicted value.

\begin{table*}[t]
\caption{Quantitative results comparing our proposed method against state-of-the-art techniques on the newly introduced video dataset. The best performing results are highlighted in bold. ``\texttt{Test}" refers to the entire test set, ``\texttt{Slow}" denotes the subset with slow-motion rate, and ``\texttt{Fast}" represents the subset with fast-motion rate. The final index is derived by averaging the results of three cross-validation experiments, with the ``$\pm$" symbol indicating the standard deviation of the errors.\label{tab:table2}}
\centering
\fontsize{9}{15}\selectfont % 设置字体大小为8pt，行距为15pt
\begin{tabular}{c|c|c|c|c|c|c|c|c}
\hline
\multicolumn{1}{c|}{}  & \multicolumn{2}{c|}{} & \multicolumn{2}{c|}{\texttt{Test}}  & \multicolumn{2}{c|}{\texttt{Slow}} & \multicolumn{2}{c}{\texttt{Fast}} \\ \hline
\multicolumn{1}{c|}{Method}  & \multicolumn{2}{c|}{Input Type}   & MAE$\downarrow$   & RMSE$\downarrow$   & MAE$\downarrow$   & RMSE$\downarrow$ & MAE$\downarrow$   & RMSE$\downarrow$ \\ \hline
CSRNet\cite{li2018csrnet} & Image & RGB  & 12.88 $\pm$ 3.1  & 17.49 $\pm$ 4.2  & 13.13 $\pm$ 3.4 & 17.57 $\pm$ 4.5 & 11.06 $\pm$ 4.4  & 15.70 $\pm$ 6.9\\ 
CAN\cite{liu2019context} & Image & RGB  & 11.36 $\pm$ 2.1 &  15.64 $\pm$ 2.4   & 11.08 $\pm$ 3.7  & 14.61 $\pm$ 4.6& 11.72 $\pm$ 4.5  & 15.03 $\pm$ 5.2\\ 
BL\cite{ma2019bayesian}  & Image & RGB  & 11.28 $\pm$ 1.9  & 15.85 $\pm$ 2.3  & 10.46 $\pm$ 3.3 & 14.22 $\pm$ 4.2 & 12.02 $\pm$ 3.9  & 15.88 $\pm$ 4.5\\ 
DMCount\cite{wang2020distribution} & Image & RGB  & 10.67 $\pm$ 1.7  & 15.01 $\pm$ 2.8 & 10.81 $\pm$ 2.9  & 14.24 $\pm$ 3.8 & 11.51 $\pm$ 2.6 & 15.32 $\pm$ 3.5\\ 
P2PNet\cite{song2021rethinking}& Image & RGB  & 10.21 $\pm$ 1.9  & 14.64 $\pm$ 2.9   & 11.11 $\pm$ 2.4  & 15.45 $\pm$ 3.4 & 9.36 $\pm$ 2.1  & 13.56 $\pm$ 2.9\\
CMCRL\cite{liu2021cross}      & Image & RGB-D & 10.25 $\pm$ 1.5 & 14.47 $\pm$ 2.8 & 10.45 $\pm$ 1.7 & 14.89 $\pm$ 3.3 & 10.02 $\pm$ 2.0 & 14.14 $\pm$ 3.4\\
GL\cite{wan2021generalized} & Image & RGB  & 9.65 $\pm$ 0.8  & 13.58 $\pm$ 1.4 & 9.16 $\pm$ 1.6 & 12.13 $\pm$ 2.1 & 10.33 $\pm$ 1.3  & 13.83 $\pm$ 2.9\\ 
EPF\cite{liu2021counting}  & Video & RGB  & 9.17 $\pm$ 0.3   & 13.10 $\pm$ 1.7   & 9.85 $\pm$ 1.5  & 12.93 $\pm$ 2.8 & 8.42 $\pm$ 1.7  & 12.23 $\pm$ 3.8\\
DEFNet\cite{zhou2022defnet} & Image & RGB-D & 10.47 $\pm$ 1.4 & 14.69 $\pm$ 2.6 & 10.24 $\pm$ 2.9 & 14.44 $\pm$ 3.6 & 10.72 $\pm$ 2.5 & 14.93 $\pm$ 3.5\\
STGN \cite{wu2022spatial}& Video & RGB  & 9.35 $\pm$ 0.6 & 13.25 $\pm$ 1.8 & 9.22 $\pm$ 1.5 & 12.24 $\pm$ 2.1 & 9.55 $\pm$ 1.1 & 13.85 $\pm$ 2.4\\
MAF\cite{ling2023motional}& Video & RGB  & 8.76 $\pm$ 0.7 & 12.43 $\pm$ 1.2 & 8.92 $\pm$ 1.8 & 12.85 $\pm$ 2.6 & 8.55 $\pm$ 1.5 & 12.05 $\pm$ 2.9 \\ \hline
Ours           & Video & RGB  & \textbf{7.80 $\pm$ 0.8} & \textbf{11.21 $\pm$ 0.9} & \textbf{8.19 $\pm$ 2.3} & \textbf{11.82 $\pm$ 3.1} & \textbf{7.23 $\pm$ 1.7} & \textbf{10.67 $\pm$ 3.1} \\ \hline

\end{tabular}
\end{table*}

\subsection{Experimental Results and Analysis}
We compared our proposed method against eleven mainstream counting models in the experiments. The compared methods include CSRNet\cite{li2018csrnet}, CAN\cite{liu2019context}, BL\cite{ma2019bayesian}, DM-Count \cite{wang2020distribution}, P2PNet\cite{song2021rethinking}, CMCRL\cite{liu2021cross}, GL\cite{wan2021generalized}, EPF\cite{liu2021counting}, DEFNet\cite{zhou2022defnet}, STGN\cite{wu2022spatial}, MAF\cite{ling2023motional}. The methods we evaluate encompass both image-based and video-based counting techniques. Specifically, the video-based approaches include EPF\cite{liu2021counting}, STGN\cite{wu2022spatial} and MAF\cite{ling2023motional}. In addition, we compare methods that use auxiliary data to assist with counting, namely CMCRL\cite{liu2021cross} and DEFNet\cite{zhou2022defnet}. Both CMCRL\cite{liu2021cross} and DEFNet \cite{zhou2022defnet} utilize not only RGB images but also depth maps generated by Depth Anything as shared inputs.

The quantitative experimental results are presented in Table \ref{tab:table2}. In addition to comprehensive evaluations on all videos within the test set, the test set has also been categorized according to motion rates to facilitate more detailed categorical assessments. In the table, ``\texttt{Test}" denotes the entire test set, ``\texttt{Slow}" indicates the test set with a slow-motion rate, and ``\texttt{Fast}" corresponds to the test set with a fast-motion rate.

The results indicate that the proposed method consistently demonstrates superior performance across all test scenarios, markedly outperforming other prevalent methods. On the entire test set, our approach achieves a MAE of $7.80$ and a RMSE of $11.21$, highlighting its stable and exceptional performance. On the \texttt{Slow} test set, the MAE is $8.19$, and the RMSE is $11.82$, while on the \texttt{Fast} test set, the MAE and RMSE are $7.23$ and $10.67$, respectively. These results clearly indicate that our method outperforms all comparative methods in both the \texttt{Slow} and \texttt{Fast} test sets, achieving the best overall performance. This further validates its adaptability to different scene complexities and its robust counting capabilities.

\begin{figure*}[t]
\centering
\includegraphics[width=1\linewidth]{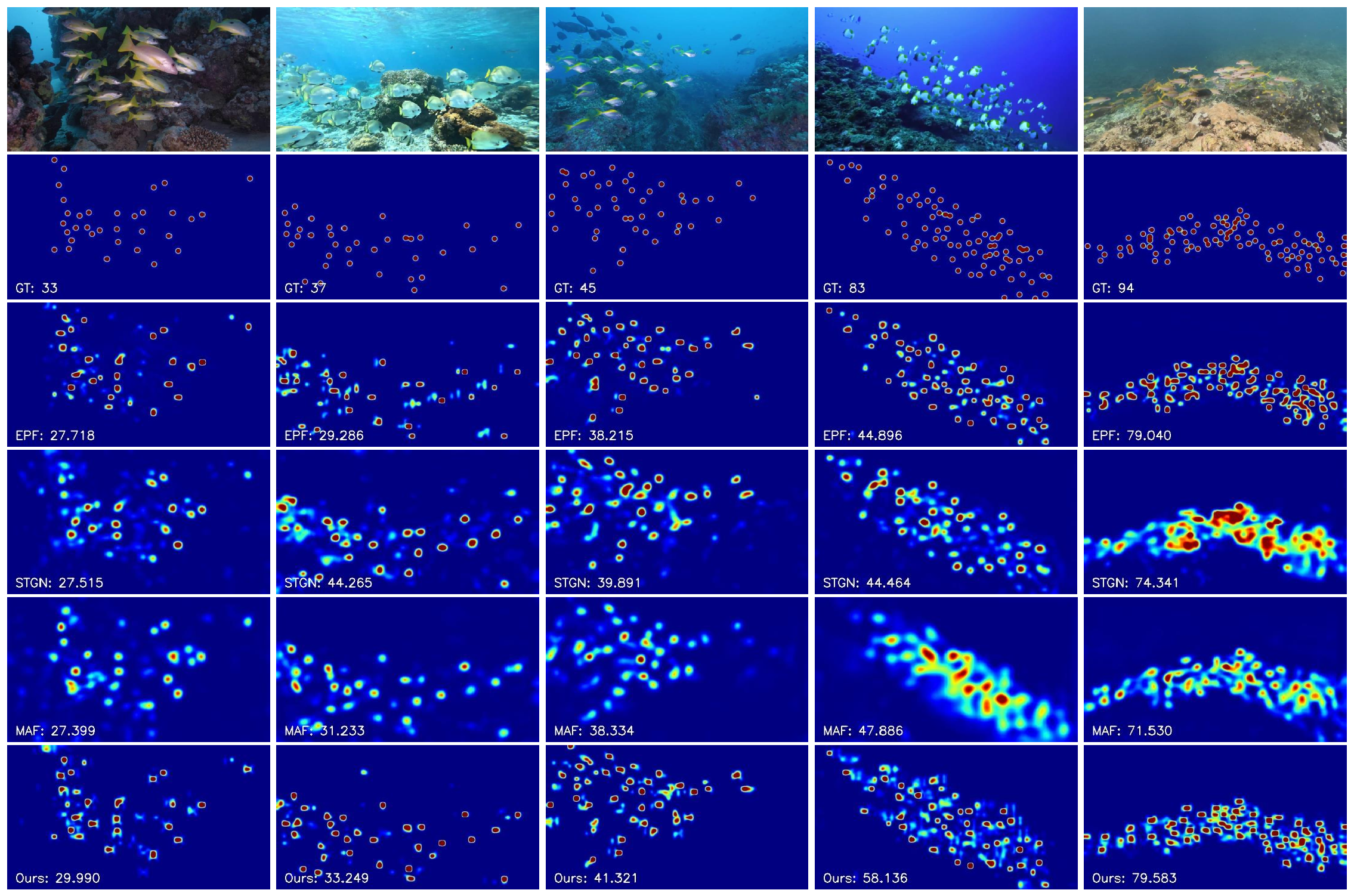}
\caption{Visualization of the counting results for EPF \cite{liu2021counting}, STGN\cite{wu2022spatial}, MAF\cite{ling2023motional} and our proposed model. The ground truth (GT) or estimated counts for each scenario are displayed in the lower left corner of the respective images. The first two rows shows the input images along with their corresponding actual density maps, while the third to sixth rows illustrate the predicted density maps generated by the aforementioned models and our proposed model, respectively.}
\label{fig:predict}
\end{figure*}

Furthermore, to provide a more intuitive understanding of our method's effectiveness across various scenes, we chose three most competitive comparison methods, i.e., EPF \cite{liu2021counting}, STGN \cite{wu2022spatial}, and MAF \cite{ling2023motional}, along with our proposed model for visual comparison. All of these methods are originally video counting models. The visualized results are presented in Fig. \ref{fig:predict}. These visualizations effectively illustrate the advantages of our approach in various video contexts, providing robust support for a comprehensive analysis of model performance.

Our method outperforms other compared mainstream methods primarily due to the following two key innovations:

• Incorporation of Depth Perception: Our approach pioneers the integration of depth perception into the task of counting indiscernible objects in underwater environments. By leveraging a joint learning mechanism for depth perception, our method not only eliminates the necessity of directly incorporating depth as input, but also enhances the spatial representation and complexity modeling of underwater scenes. This enables the model to achieve robust and accurate performance in scenarios characterized by severe occlusion, challenging lighting conditions, and diverse shapes of objects.

• Multi-Scale Perception and Adaptive Fusion: We employ a multi-steam dilated convolutional architecture to capture features across various scales. Additionally, we integrate a motion-weighting mechanism that incorporates optical flow information into the network, adaptively emphasizing scale-differentiated features. This approach effectively mitigates the challenges posed by motion scale variations due to differing speeds, thereby ensuring high counting accuracy even in complex underwater environments.

\subsection{Ablation Study}
\subsubsection{Ablation Study on Key Modules} 
To demonstrate the effectiveness of the core components in our network, we conducted a series of ablation studies designed to systematically evaluate the contributions of the depth-enhanced encoder and the adaptive flow estimation module. By progressively removing or modifying these components, we were able to assess their individual impact on overall performance. The experimental results, summarized in Table \ref{tab:table3}, clearly illustrate the performance of the model in different configurations.

\begin{table}[t]
\centering
\fontsize{10}{15}\selectfont % 设置字体大小为10pt，行距为15pt
\caption{The results of the ablation study on the depth-enhanced encoder (DEE), motion enhancement module (MEM) and adaptive flow feature fusion (AFFF), respectively. All experiments were carried out on the proposed dataset. \label{tab:table3}}
\begin{tabular}{c|c|c|c|c} 
\hline
\multicolumn{1}{c|}{DEE} & MEM & AFFF & MAE$\downarrow$ & RMSE$\downarrow$ \\ \hline
             &   &   & 9.17 $\pm$ 0.3   & 13.10 $\pm$ 1.7 \\ 
\checkmark  &  &    & 8.62 $\pm$ 0.5   & 12.65 $\pm$ 1.1  \\ 
           & \checkmark  & \checkmark  & 8.09 $\pm$ 0.7 & 11.58 $\pm$ 0.9 \\ 
\checkmark & \checkmark & \checkmark & \textbf{7.80 $\pm$ 0.8} & \textbf{11.21 $\pm$ 0.9} \\ \hline
\end{tabular}
\label{tab:xiao1}
\end{table}

\begin{figure*}[t]
\centering
\includegraphics[width=1\linewidth]{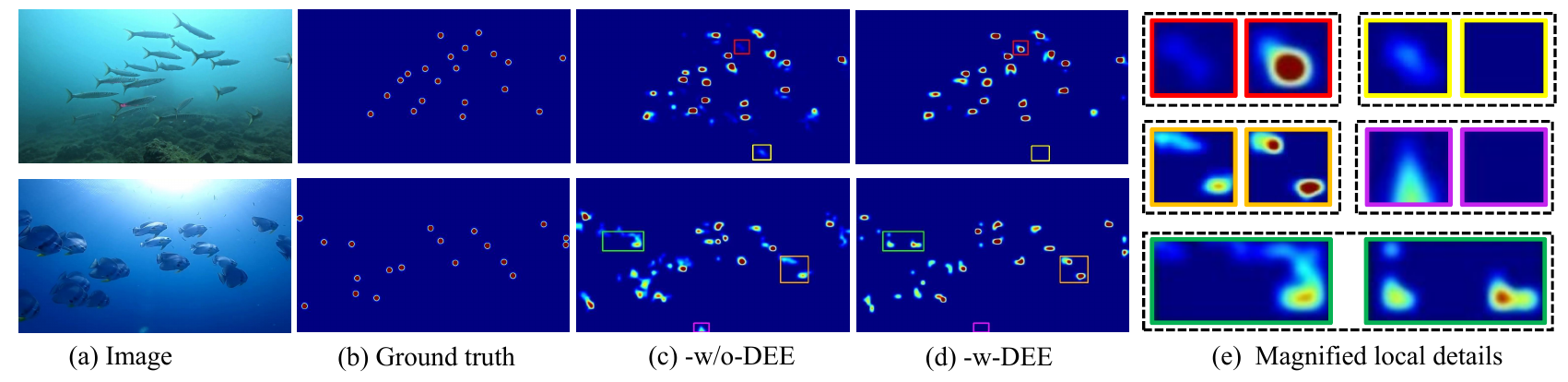}
\caption{Visualized results of the ablation study on the Depth-Enhanced Encoder (DEE). (c) and (d) show the results without and with the DEE, respectively. (e) provides a detailed comparison of five local regions, which more clearly demonstrates the improvement brought by the DEE. The yellow and pink boxes highlight that foreground objects are incorrectly detected in the background without the DEE, while the DEE significantly reduces these false detections. The other three boxes reveal that the model fails to correctly predict foreground objects without the DEE, but successfully predicts them with the DEE.}
\label{fig:Visual_DEE}
\end{figure*}

As illustrated in Table \ref{tab:table3}, the incorporation of depth information via the depth-enhanced encoder leads to a substantial improvement in density estimation, with reductions of $6.0\%$ in MAE and $3.4\%$ in RMSE. Fig. \ref{fig:Visual_DEE} shows the effectiveness of depth-aware features (depth-enhanced encoder) in improving density estimation. We compared the performance of the baseline model against that of the model incorporating the depth-enhanced encoder. The baseline model exhibited a significant number of false detections, including failures to predict foreground objects and erroneous classifications of background elements as foreground. By integrating depth features, the depth-enhanced encoder mitigates these errors, thereby improving the model's ability to distinguish between foreground and background elements and enhancing foreground feature extraction. This improvement is particularly evident in the boxed regions highlighted in Fig. \ref{fig:Visual_DEE}.
 
\begin{figure}[t]
\centering
\includegraphics[width=1\linewidth]{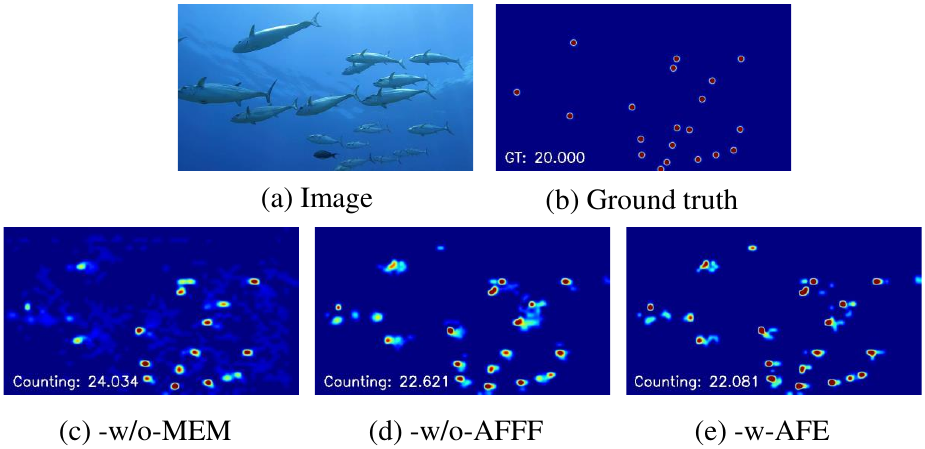}
\caption{Visualized comparison of the ablation study on MEM, AFFF and AFE modules. (c) -w/o-MEM: without motion enhancement module; (d) -w/o-AFFF: without adaptive flow feature fusion; (e) -w-AFE: with adaptive flow estimation, which integrates both motion enhanced module and adaptive flow feature fusion.}
\label{fig:Visual_AFE}
\end{figure}

Furthermore, the method that incorporates adaptive flow estimation also showed significant improvements, reducing MAE and RMSE by $11.8$\% and $11.6$\%, respectively. Fig. \ref{fig:Visual_AFE} illustrates the significant benefits of the adaptive flow estimation module. The motion enhancement module is capable of effectively capturing the motion features of indiscernible objects, thereby ensuring accurate tracking and counting of object movements in complex scenes, particularly under conditions of significant background interference. In addition, the motion weight-guided adaptive fusion further improves the performance of the model. Through adaptive adjustment of these weights, the module can more flexibly and accurately capture multi-scale perception features.

\begin{table}[t]
\centering
\fontsize{10}{15}\selectfont % 设置字体大小，行距
\caption{Experimental comparison between single-stream and multi-stream convolutional layer structures. The single-stream convolutional layer structure comprises three distinct configurations of convolution layers. The multi-stream convolutional layer structure is further categorized into fixed-weight fusion and adaptive motion weight-guided fusion.\label{tab:table4}}
\begin{tabular}{c|c|c|c} 
\hline
\multicolumn{1}{c|}{MEM} &   & MAE$\downarrow$ & RMSE$\downarrow$ \\ \hline
$\times$     & $3 \times 3$ Conv, rate = $1$  & 9.48 $\pm$ 0.3  & 13.92 $\pm$ 1.2  \\ 
$\times$     & $3 \times 3$ Conv, rate = $2$   & 9.21 $\pm$ 0.4 & 13.17 $\pm$ 1.0 \\ 
$\times$     & $3 \times 3$ Conv, rate = $3$     & 9.76 $\pm$ 0.6 & 14.08 $\pm$ 1.5 \\ \hline
\checkmark  & Fixed weights     & 8.34 $\pm$ 0.8 & 12.61 $\pm$ 1.1 \\            
\checkmark  & Adaptive weights  & \textbf{8.09 $\pm$ 0.7} & \textbf{11.58 $\pm$ 0.9} \\ \hline
\end{tabular}
\end{table}

\begin{figure*}[t]
\centering
\includegraphics[width=0.98\linewidth]{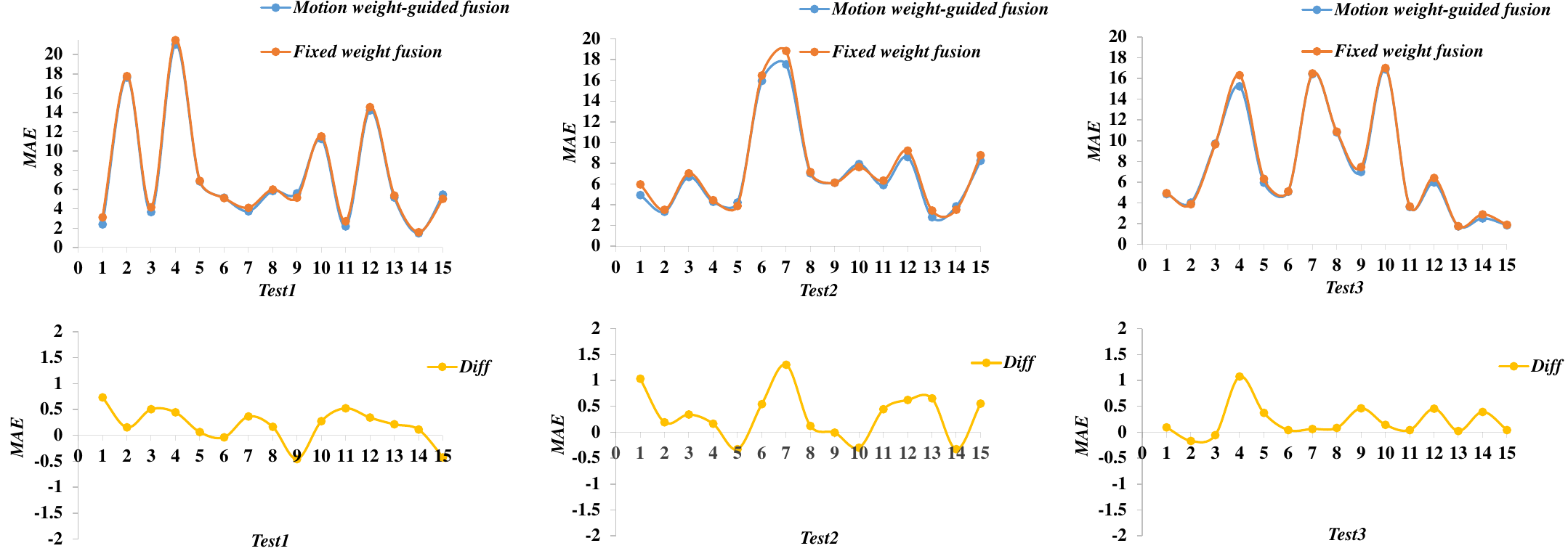}
\caption{Comparison of performance metrics (in terms of MAE) and analysis of differences between motion weight-guided fusion and fixed weight fusion. The first row illustrates the MAE comparison between the two fusion methods, whereas the second row highlights the performance discrepancies.}
\label{fig:Diff}
\end{figure*}

\subsubsection{Ablation Study on Motion Weight-Guided Fusion}\label{sec:ablaMWGF}
We further conduct an ablation experiment to evaluate the effectiveness of using adaptive motion weight-guided fusion compared to fixed-weight fusion. In this experiment, we compare the performance of a single-stream dilated convolutional structure with that of a multi-stream dilated convolutional architecture. The multi-stream dilated convolutional architecture is further divided into two settings: fixed-weight fusion and adaptive motion weight-guided fusion. The experimental results are summarized in Table \ref{tab:table4}. This comparative analysis systematically evaluates the impact of different structures and fusion methods on model performance, validating the benefits of adaptive motion weights.

The fixed fusion weights are uniformly assigned as $1/3$ for each component. Here, ``$3 \times 3$ Conv, rate = $1$" denotes a single-stream dilated convolutional structure utilizing a $3 \times 3$ convolution kernel. Meanwhile, ``$3 \times 3$ Conv, rate = $2$" and ``$3 \times 3$ Conv, rate = $3$" indicate single-stream dilated convolutional structures that employ $3 \times 3$ dilated convolution kernels with dilation rates of $2$ and $3$, respectively. We evaluated the performance of various convolutional architectures within the model. The results indicate that the adaptive fusion method employing motion-guided weights outperforms alternative approaches, validating its superiority. This method captures dynamic information more accurately through the use of motion-guided weights and substantially enhances the model's robustness in handling complex backgrounds and variations in target motion scales.

To more clearly demonstrate the superiority of using motion weights for adaptive fusion over fixed weight settings, we compare the MAE metrics for each test video in the three test sets, as shown in Fig. \ref{fig:Diff}. The first row of the figure presents the MAE comparison between the two fusion methods, while the second row shows the differences. A positive difference indicates that the method using motion weight-guided fusion outperforms fixed weight fusion, whereas a negative difference suggests inferior performance. This allows for a more detailed observation of the impact of each fusion method on individual test videos.

\subsection{Crowd Experimental Results and Analysis}

\begin{table*}[t]
\caption{Quantitative comparisons of our results with state-of-the-art methods on three crowd video datasets are presented. The top and second-best performing results are highlighted in \textcolor{red}{\textbf{red}} and \textcolor{blue}{\textbf{blue}} fonts, respectively.\label{tab:table5}}
\centering
\fontsize{10}{15}\selectfont % 设置字体大小为8pt，行距为15pt
\begin{tabular}{c|c|c|c|c|c|c|c}
\hline
\multicolumn{2}{c|}{Datasets}  & \multicolumn{2}{c|}{\texttt{Classroom}}  & \multicolumn{2}{c|}{\texttt{Mall}}  & \multicolumn{2}{c}{\texttt{FDST}}\\ \hline
\multicolumn{1}{c|}{Method}  & Type      & MAE$\downarrow$   & RMSE$\downarrow$ & MAE$\downarrow$   & RMSE$\downarrow$ & MAE$\downarrow$   & RMSE$\downarrow$ \\ \hline
CSRNet\cite{li2018csrnet}   & Image     & 3.42  & 3.86 & 1.84  & 2.34 & 1.98  &2.51 \\ 
BL \cite{ma2019bayesian}       & Image     & 2.86   & 3.14 & 1.96  & 2.49 & -  & - \\ 
DMCount \cite{wang2020distribution} & Image   & 2.86   &3.94 & 2.02  & 2.52 & -  & - \\ 
LSTN \cite{fang2019locality}   & Video   & 3.02   & 3.90 & 2.03  & 2.60 & 3.35  & 4.45\\ 
EPF \cite{liu2021counting}  & Video    & 1.68  & 2.06 & 2.06   &2.61   & 2.10   & 2.46  \\
MAF \cite{ling2023motional}  & Video     & \textcolor{blue}{\textbf{1.64}}  & \textcolor{blue}{\textbf{2.00}} & \textcolor{red}{\textbf{1.51}}  & \textcolor{red}{\textbf{1.93}} & \textcolor{red}{\textbf{1.19}}  & \textcolor{red}{\textbf{1.56}}\\ \hline 
Ours                   & Video  & \textcolor{red}{\textbf{1.54}}  & \textcolor{red}{\textbf{1.83}} & \textcolor{blue}{\textbf{1.75}}  & \textcolor{blue}{\textbf{2.22}}   & \textcolor{blue}{\textbf{1.81}}   & \textcolor{blue}{\textbf{2.35}}  \\ \hline
\end{tabular}
\end{table*}

To verify the generality of our model, we conduct experiments on three different crowd video datasets: \texttt{Classroom}\cite{ling2023motional}, \texttt{Mall}\cite{chen2012feature}, and \texttt{FDST}\cite{fang2019locality}.

The \texttt{Classroom} dataset \cite{ling2023motional} was developed by recording videos in a classroom environment. The videos were captured with a frame interval of $10$ frames, resulting in a total of $1,290$ frames at a resolution of $1280 \times 720$ pixels. The number of individuals present in each frame varies considerably, ranging from a minimum of $1$ to a maximum of $51$, with an average of approximately $30.3$ individuals per frame. Keeping with the methodology described in \cite{ling2023motional}, the dataset has been divided into two equal parts, where the first half is designated for training purposes and the second half for testing.

The \texttt{Mall} dataset \cite{chen2012feature} was collected in a shopping mall equipped with surveillance cameras. This dataset captures both moving and stationary pedestrians under complex lighting conditions. It comprises $2,000$ annotated video frames, each with a resolution of $640 \times 480$ pixels and a frame rate of less than $2$ FPS. In accordance with the experimental setup described by Chen et al. \cite{chen2012feature}, the first $800$ frames are designated for training purposes, while frames $801$ to $2,000$ are reserved for testing.

\begin{figure}[t]
\centering
\includegraphics[width=1\linewidth]{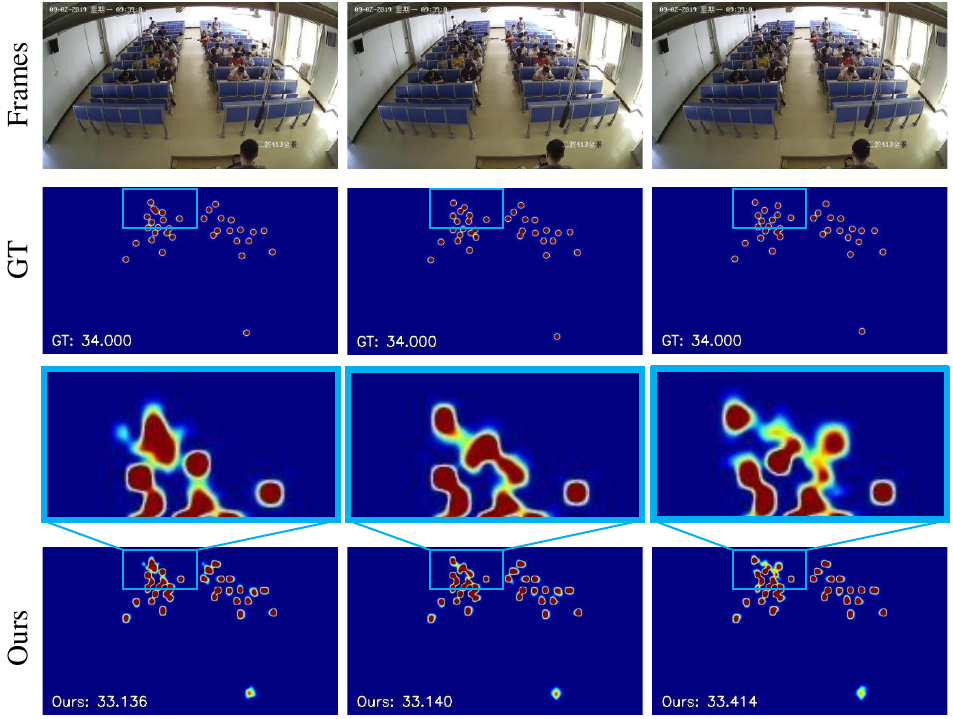}
\caption{Visualization of our model across three consecutive frames within the \texttt{Classroom} test set, highlighting the region of interest where the most significant changes in crowd dynamics are observed.}
\label{fig:Visual_class}
\end{figure}

\begin{figure}[t]
\centering
\includegraphics[width=1\linewidth]{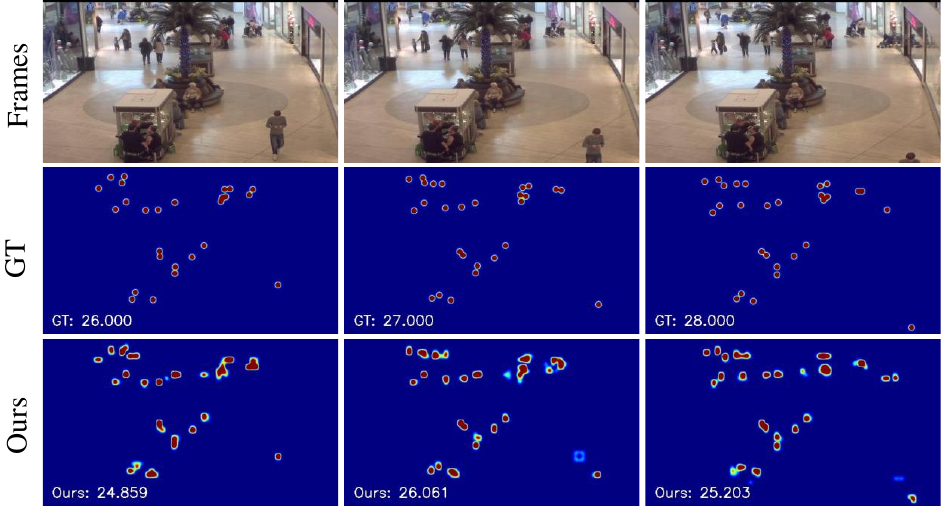}
\caption{Visualization of the performance of our model across three consecutive frames on the \texttt{Mall} test set.}
\label{fig:Visual_Mall}
\end{figure}

The \texttt{FDST} dataset \cite{fang2019locality} is presently recognized as the largest and most comprehensive crowd video dataset, encompassing $100$ videos across $13$ diverse scenes in both indoor and outdoor settings. Each video consists of $150$ frames and is available in resolutions of either $1920 \times 1080$ or $1280 \times 720$. Adhering to the configurations outlined by \cite{fang2019locality}, a total of $60$ videos are designated for training purposes, while the remaining $40$ videos are reserved for testing.

\begin{figure*}[t]
\centering
\includegraphics[width=1\linewidth]{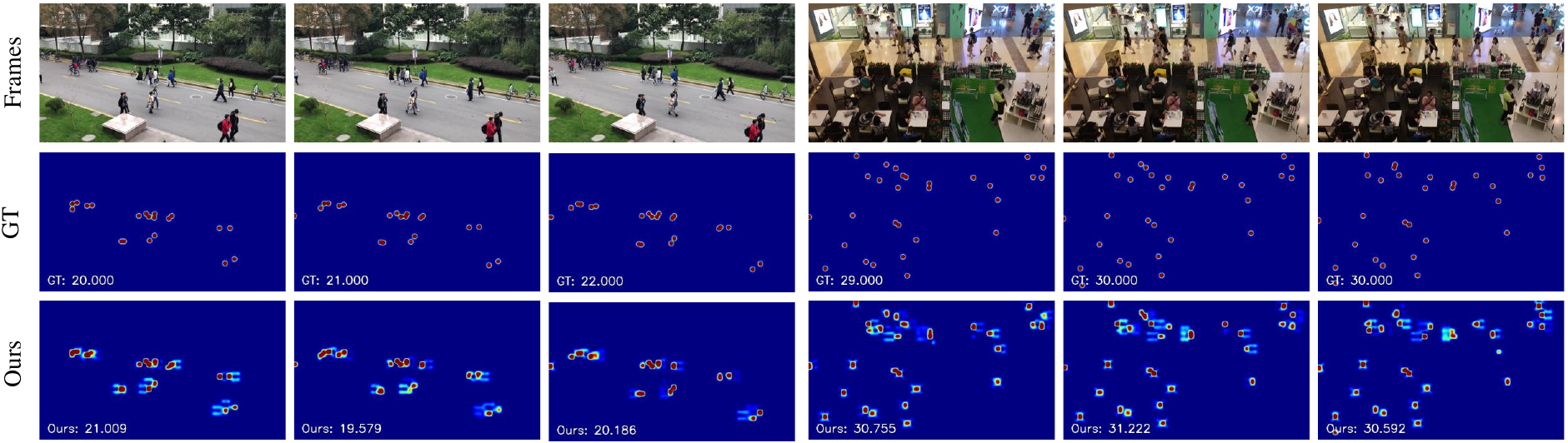}
\caption{Visualization of the performance of our model across three consecutive frames in both outdoor and indoor settings on the \texttt{FDST} test set.}
\label{fig:Visual_FDST}
\end{figure*}

We evaluate our proposed method against three image-based and three video-based crowd counting approaches in three crowd video datasets, with the results summarized in Table \ref{tab:table5}. As illustrated, our method demonstrates superior performance on the \texttt{Classroom} dataset \cite{ling2023motional}, achieving the lowest MAE and RMSE. Specifically, compared to the second-best performing method, MAF \cite{ling2023motional}, our approach reduces MAE by $6.1$\% and RMSE by $8.5$\%. These results indicate that the integration of depth perception alleviates the challenges associated with poor lighting and severe occlusion in indoor environments. Fig. \ref{fig:Visual_class} presents a visual representation of the results of our method, displaying three consecutive frames that emphasize dynamic crowd changes, with a particular focus on the most salient regions. Despite movement and variability within the crowd, our method consistently captures these significant dynamic transitions accurately. 

Our method does not perform as well on the \texttt{Mall} and \texttt{FDST} datasets as the MAF model \cite{ling2023motional}. MAF \cite{ling2023motional} extracts motion features through bidirectional frame difference and combines these motion features with static features of the current frame for crowd counting. When processing the \texttt{Mall} and \texttt{FDST} datasets, the bidirectional frame difference can more easily eliminate the effects of disturbances because the background and lighting conditions are relatively stable. While, our method still demonstrated notable improvements compared to the EPF model \cite{liu2021counting}. To visually demonstrate the efficacy of our approach, Fig. \ref{fig:Visual_Mall} illustrates the visualization results on the \texttt{Mall} dataset \cite{chen2012feature}, while Fig. \ref{fig:Visual_FDST} provides the visualization outcomes for both indoor and outdoor scenes from the \texttt{FDST} dataset \cite{fang2019locality}. These visualizations serve to further substantiate the adaptability and robust performance of our method in various environments.

\subsection{Discussion}
\begin{table}[t]
\centering
\fontsize{10}{15}\selectfont % 设置字体大小为8pt，行距为15pt
\caption{Efficiency analysis of the compared methods. Ours-w/o-RAFT : without the pre-training model of RAFT. \label{tab:table6}}
\begin{tabular}{c|c|c|c} 
\hline
\multicolumn{1}{c|}{Method} & Type   & Parameters(M)  & FPS \\ \hline
CSRNet \cite{li2018csrnet}    &Image   & 68.00   &  $ > $30    \\ 
CAN \cite{liu2019context}     &Image     & 75.02    &  $ > $30 \\ 
EPF \cite{liu2021counting}     &Video     & 78.10   & 22 $\sim$ 23 \\ 
STGN  \cite{wu2022spatial}   &Video     & 47.35   &  $ > $30 \\ 
MAF  \cite{ling2023motional}    &Video    & 250.35    & 26 $\sim$ 27 \\ \hline
Ours-w/o-RAFT  &Video     & 97.50   & 21 $\sim$ 22 \\ 
Ours   &Video   & 118.63  & 8 $\sim$ 9   \\   \hline
\end{tabular}
\label{tab:efficiency}
\end{table}
\subsubsection{Efficiency Analysis} In practical applications, both model size and operational efficiency serve as critical metrics for evaluating performance. In the Table \ref{tab:table6}, we summarized the operation efficiency and model size of some comparison methods. The experiment performed operations on images with a size of $640 \times 360$. Generally, the single-image-based approach can meet the real-time requirements. The frame rate of our method can reach more than $20$ frames without calculating the pre-training model of RAFT \cite{teed2020raft}. However, if the pre-training model of RAFT \cite{teed2020raft} is loaded together with our method, the frame rate is about $8\sim9$ frames. Given that in practical counting tasks it is not essential to generate predictions for each frame, the current processing speed is sufficient to meet the requirements of real-world applications. The frame rate of MAF \cite{ling2023motional} approaches real-time; however, it has the largest number of model parameters, making deployment on edge computing devices challenging. In contrast, while STGN \cite{wu2022spatial} exhibits inferior performance compared to other video counting methods, its smaller number of model parameters and a frame rate exceeding $30$ FPS facilitate easier deployment on edge computing devices.
 
\subsubsection{Perspectives of Future Works and Applications} 
The limited visibility in underwater environments, combined with the high similarity between foreground and background in terms of appearance, color, and texture, poses significant challenges for distinguishing and accurately counting foreground objects. Consequently, effective differentiation between foreground and background is crucial for addressing the issue of indiscernible object counting in underwater scenes. Future research should explore the integration of additional information, such as semantics, to enhance feature extraction in regions with indiscernible targets. It holds promise for improving model performance and operational efficiency, thereby facilitating more challenging and complicated downstream applications in real-world scenarios.

\section{conclusion}
Indiscernible object counting in marine environments has extensive applications in marine resource surveys, ecological monitoring, and fisheries management. Image-based counting methods exhibit notable limitations in capturing spatial and temporal scale information, hindering their ability to satisfy the requirements of practical applications. In contrast, video-based counting is emerging as a promising approach due to its potential capability to address these challenges. In this paper, we propose VIMOC-Net, a depth-assisted network with adaptive motion-differentiated feature encoding for indiscernible marine object video counting. It leverages depth-aware features to enhance the spatial feature representation of indiscernible objects. Meanwhile, adaptive flow estimation on multi-scale perception features is conducted effectively addressing the issue of motion scale variations. The performance of VIMOC-Net has been rigorously validated on both our self-built challenging VIMOC dataset, which comprises $50$ videos with approximately $40,800$ annotated points, and several widely recognized video crowd counting datasets. Specifically, when evaluated on the VIMOC dataset, VIMOC-Net achieved a reduction in error estimation levels by $10.96$\% in MAE and $9.8$\% in RMSE compared to current state-of-the-art methods. In addition, ablation studies highlight the significant contributions of the proposed modules to overall performance.

\ifCLASSOPTIONcaptionsoff
  \newpage
\fi

\bibliographystyle{IEEEtran}

\bibliography{reference}


\end{document}